\documentclass[letterpaper, 10 pt, conference]{ieeeconf}  

\IEEEoverridecommandlockouts                              

\usepackage{times}

\usepackage{multicol}
\usepackage[bookmarks=true]{hyperref}

\usepackage[utf8]{inputenc} 
\usepackage[T1]{fontenc}    
\usepackage{hyperref}       
\usepackage{url}            
\usepackage{booktabs}       
\usepackage{amsfonts}       
\usepackage{nicefrac}  
\usepackage{xcolor}
\usepackage{microtype}      
\usepackage{lipsum}
\usepackage{graphicx}
\usepackage{subfigure}
\usepackage{multicol}  
\usepackage{multirow}
\usepackage{stfloats}
\usepackage{textcomp}

\usepackage{amsmath,amssymb,amsfonts}
\usepackage{algorithmic}
\usepackage{textcomp}
\usepackage{xcolor}
\usepackage{algorithm}
\usepackage{multicol}  
\usepackage{multirow} 
\usepackage{subfigure}
\usepackage{dashrule}

\usepackage{listings}
\usepackage{listings}
\usepackage{xcolor}
\usepackage{pdfpages}

\definecolor{codegreen}{rgb}{0,0.6,0}
\definecolor{codegray}{rgb}{0.5,0.5,0.5}
\definecolor{codepurple}{rgb}{0.58,0,0.82}
\definecolor{backcolour}{rgb}{0.95,0.95,0.92}

\lstdefinestyle{python}{
    backgroundcolor=\color{backcolour},   
    commentstyle=\color{codegreen},
    keywordstyle=\color{magenta},
    numberstyle=\tiny\color{codegray},
    stringstyle=\color{codepurple},
    basicstyle=\ttfamily\footnotesize,
    breakatwhitespace=false,         
    breaklines=true,                 
    captionpos=b,                    
    keepspaces=true,                 
    numbers=left,                    
    numbersep=5pt,                  
    showspaces=false,                
    showstringspaces=false,
    showtabs=false,                  
    tabsize=2
}

\begin{document}

\title{LLM Granularity for On-the-Fly Robot Control}


\author{Peng Wang, Mattia Robbiani, Zhihao Guo
\thanks{Department of Computing and Mathematics, Manchester Metropolitan  University, Manchester, M15 6BH, UK. 
        {\tt\small Contact: p.wang@mmu.ac.uk}}%
}

\maketitle

\begin{abstract}

Assistive robots have attracted significant attention due to their potential to enhance the quality of life for vulnerable individuals like the elderly. The convergence of computer vision, large language models, and robotics has introduced the `visuolinguomotor' mode for assistive robots, where visuals and linguistics are incorporated into assistive robots to enable proactive and interactive assistance. This raises the question: \textit{In circumstances where visuals become unreliable or unavailable, can we rely solely on language to control robots, i.e., the viability of the `linguomotor` mode for assistive robots?} This work takes the initial steps to answer this question by: 1) evaluating the responses of assistive robots to language prompts of varying granularities; and 2) exploring the necessity and feasibility of controlling the robot on-the-fly. We have designed and conducted experiments on a Sawyer cobot to support our arguments. A Turtlebot robot case is designed to demonstrate the adaptation of the solution to scenarios where assistive robots need to maneuver to assist. Codes will be released on GitHub soon to benefit the community.

\end{abstract}

\IEEEpeerreviewmaketitle

\section{Introduction}

\textbf{RQ1}: \textit{What are the advantages and disadvantages of using qualitative language prompts in an assistive robot setup? }

Natural language has long been regarded as a fundamental capability for humans to communicate and interact with their environments~\cite{shi2024can,carr2023human}. 
Enabling language control of assistive robots will create an interactive yet natural way of assistance providing and help to improve the life quality for individuals with motor impairments.
The recent success of foundational models such as Large Language Models (LLMs) has catalyzed the convergence of robotics, computer vision, and LLMs, making nuanced robot assistance promising. For instance, \cite{padmanabha2024voicepilot} presents VoicePilot, a framework for integrating LLMs as speech interfaces for physically assistive robots. The framework aims to enhance the well-being and independence of individuals with motor impairments by enabling natural and effective communication with robots. 
However, LLMs could misinterpret prompts like `\textit{scrape down the yogurt bowl and then feed me some}', leading to discomfort by putting the scoop too deep into the mouth. Default depth and speed values were introduced to deal with the issue, which limited the flexibility for robot assistance providing. Another issue noted is that prompts are provided to LLMs to generate Python codes that are deployed on the robot subsequently, which hinders on-the-fly control and poses risks if the robot malfunctions.

To provide an assistive robot that considers human feelings and capabilities, \cite{cao2024smart} utilized LLMs to help generate actions for the robot by taking robot state observations as prompts. The dependence on the state observations undermines the generality of the solution as they are not always accessible.  
 In parallel, we also see the emergence of Visual Language Models (VLMs) and their applications in assistive tasks. For instance, \cite{patel2023pretrained} proposed a VLM-based planner to assist humans in task completion. Users can specify a goal in qualitative prompts like `\textit{build a shelf}', and visual history will be next used to generate actions to achieve the goal. 

We notice that most prompts used in these works are qualitative like `\textit{move lower'}~\cite {cao2024smart,patel2023pretrained,shi2024yell} without quantifying how low is `lower'. 
The ambiguities in such prompts spark whether we can rely on qualitative prompts in assistive robotics where safety is paramount. Recent solutions have focused on using LLMs to provide real-time (online) corrections~\cite{shi2024yell} or post-task (offline)~\cite{cui2023no} corrections to deal with language ambiguities. In these cases, humans observe the mistakes made by robots and provide feedback to either the robot or the training data. The correction data is then used to fine-tune the model and improve its performance. However, these models currently struggle to generalize from one case to another, and the restriction of language commands used to train the models limits their generality and scalability.

\begin{figure*}[t!]
    \centering
    \includegraphics[width=0.90\linewidth]{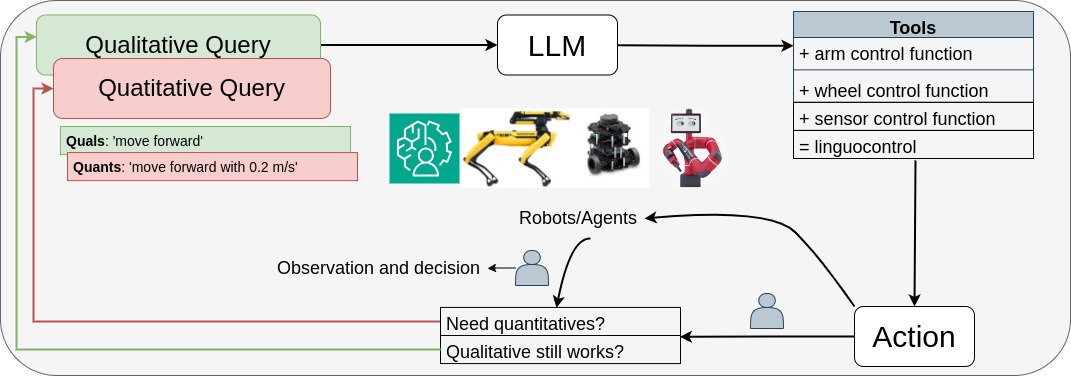}
    \caption{The pipeline to test how language granularity affects robot control performance. Humans will act as the observer and the decision-maker. Humans will use language prompts of different granularity to control the robot and change the granularity based on observations.}
    \label{fig:qualquatflow}
\end{figure*}

\textbf{RQ2}: \textit{Is it necessary to control an assistive robot on-the-fly using language?}

Controlling an assistive robot on-the-fly using language becomes apparent in dynamic and unpredictable environments~\cite{broad2016towards}. \cite{shi2024can} states that `forming robot policies using LLMs for spontaneous tasks' is one of the promising trends where on-the-fly is needed to deal with unpredictability. Pre-programmed or even learned commands could fall short in such scenarios, making real-time language control essential for adaptation.  We also reckon that on-the-fly language control is vital for effective human-robot interactive assistance providing. In settings where long-horizon interactions are needed to assist, swift responses to language instructions are necessary to maintain efficient workflow and cooperation~\cite{carr2023human}. For instance, \cite{patel2023pretrained,lynch2023interactive} have trained LLMs or VLMs on a tremendous amount of datasets to achieve long-horizon assistive tasks such as `\textit{make a smiley face out of blocks}' or `\textit{make a coffee}'. The challenge is that model performance is only evaluated on whether the task is completed. In an assistive robot setting, we need to quantify the success of a model to a finer level, e.g., \textit{has put a glass on the very edge of a table} and \textit{has put a glass on the center of a table} can both be regarded as successful in carrying out the task described as `\textit{put a glass on the table}'. The latter, however, should be a `more' pleasant outcome than the former.

Enabling on-the-fly control of robots using language could compensate for the `unpleasant' results caused by ambiguities in the qualitative language and henceforth, enable more effective robotic assistance to humans.

\begin{figure}[t!]
    \centering
    \includegraphics[width=0.90\linewidth]{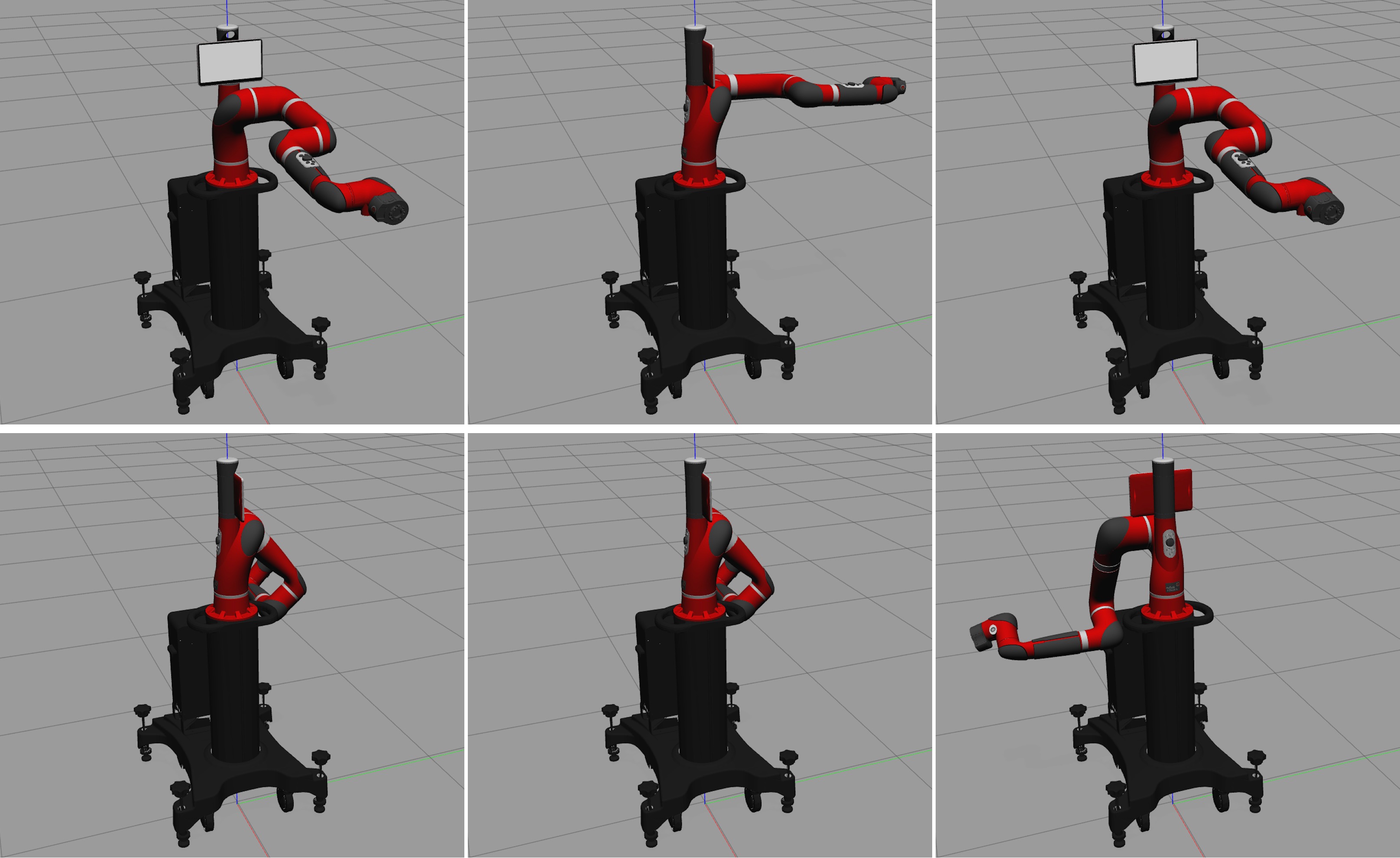}
    \caption{The poses of the robots after executing the six prompts. From top left to bottom right: prompts \textcircled{1} to \textcircled{6} in the first column of Table \ref{tab:sayerdata}, respectively.}
    \label{fig:sawyerfigure}
\end{figure}

\begin{table*}[ht!]
    \centering
    \begin{tabular}{l|c|c|c|c|c|c|c}
        \hline
        \textbf{Prompts} & {right\_j0} & {right\_j1} & {right\_j2} & {right\_j3} & {right\_j4} & {right\_j5} & {right\_j6} \\ \hline \hline
        \textcircled{1} move the arm up & ~0 & 0.068 & 0.014 & 0.020 & -0.007 & 0.007 & 0 \\ \hline
        \textcircled{2} rotate the base 90 degrees & 1.57 & 0.068 & 0.014 & 0.020 & -0.007 & 0.007 & 0 \\ \hline
        \textcircled{3} move all joints to 0 & 0 & 0.068 & 0.014 & 0.020 & -0.007 & 0.007 & 0 \\ \hline
        \textcircled{4} move right\_j3 by 90 degrees & -0.073 & 1.148 & 1.603 & 1.565 & -0.095 & -0.432 & -0.732 \\ \hline
        \textcircled{5} move right\_j0 by 90 degrees & 1.481 & 0.851 & 1.543 & 1.538 & -0.094 & -0.433 & -0.723 \\ \hline
        \textcircled{6} move right\_j0 to the left by 180 degrees & -3.050 & 1.093 & 1.649 & 1.535 & -0.094 & -0.434 & -0.073 \\ \hline
    \end{tabular}
    \caption{Prompts and robot states.}
    \label{tab:sayerdata}
\end{table*}

\section{Approaches}
To answer the two questions, we have designed two sets of experiments using a Sawyer cobot arm and another set with the TurtleBot mobile robot. We use ROS noetic and Ubuntu 20.04 LTS to control the robots. The Groq API with LLAMA3 70B LLM model is used for language inference. It is worth noting that even though we used the two robots, the whole pipeline is robot-agnostic as long as the robot is compatible with ROS. No fine-tuning of LLMs was applied. We used tools (a.k.a. function calls) to automatically wrap and publish the response from LLM to the corresponding robot control topic. In such a manner, we can achieve end-to-end linguomotor control. The whole pipeline can be found in Fig. \ref{fig:qualquatflow}. Note that human users can be regarded as observers and decision-makers, and the prompts will pass decisions to the robots.

To investigate how language granularity affects robot control performance on the fly, we have designed prompts ranging from qualitative to quantitative, as illustrated in Table \ref{tab:sayerdata}. Qualitative prompts do not specify values such as speed or duration, while quantitative prompts include these specifics. Due to space constraints, we have provided the prompts, results, and analysis for the Sawyer cobot only experiment in this section. The results of the Sawyer Pick-and-Place and the TurtleBot mobile robot experiments are available in the Appendix.


\section{Results and Discussions}

Table \ref{tab:sayerdata} and Figure \ref{fig:sawyerfigure} present the results for the Sawyer Cobot. We observed that when qualitative prompts, such as `\textit{move the arm up}' are used, LLM generates small joint movement values, which are then executed by the cobot. The states of the cobot before and after executing the prompt are detailed in one example in Section~\ref{subsec:sawyermsg}. The small movements are reasonable as they help avoid collisions with objects and humans, which is critical to ensure safety in assistance. However, this conservative response necessitates additional prompts to move the robot to the desired poses. 

Humans (or available visual) can typically estimate the distance the cobot needs to move, allowing us to reduce the ambiguities in language by specifying precise actions like `\textit{rotate the base 90 degrees}' or even more complex commands such as `\textit{move right\_j0 to the left by 180 degrees}'. Table \ref{tab:sayerdata} shows the results of these quantified language prompts, demonstrating that the cobot can execute the commands and reach the expected poses. All results are obtained by controlling the cobot on the fly. Figure \ref{fig:sawyerfigure} further illustrates that quantitative language can enable cobots to complete more complex tasks effectively.

\section{Conclusion} 
\label{sec:conclusion}
This paper initiates the validation of the `linguomotor` mode by evaluating robot responses to human language commands of different granularities in assistive robot contexts. Through experiments conducted on a Sawyer cobot and a Turtlebot (mobile assitance), we provide foundational insights into the feasibility of relying solely on language for robotic control in various scenarios on-the-fly.
\section*{APPENDIX}



\subsection{Results for TurtleBot mobile robot}\label{subsec:promptturtle}

\begin{table*}[hb!]
    \centering
    \begin{tabular}{l|c|c|c|c|c|c}
        \hline
        \textbf{Prompts} & {$\textbf{v}_x(\text{m/s})$} & {$\omega(\text{degree/s})$} & {$t(\text{s})$} & \textbf{\( x \)} & \textbf{ \( y \)} & $\theta$ (degree) \\ \hline \hline
        initial pose & n/a & n/a & n/a & 0.0 & 0.0 & 0.0 \\ \hline
        \textcircled{1}move forward & 0.5 & 0.0 & 1.0 & 0.088 & 0 & -0.085 \\ \hline
        \textcircled{2}move back & 0.0 & 0.0 & 1.0 & 0.263 & 0 & -1.066 \\ \hline
        \textcircled{3}move along x-axis with a speed of 0.05 m/s for 5 seconds
& 0.05 & 0.0 & 5.0 & 0.470 & -0.005 & -1.312 \\ \hline
        \textcircled{4}move backward for 2 seconds& -1.0 & 0.0 & 2.0 & -0.258 & -0.004 & -0.007 \\ \hline
        \textcircled{5}move forward at a speed of 0.8 for 2 seconds & 0.8 & 0.0 & 2.0 & -0.872 & -0.042 & -22,0 \\ \hline
        \textcircled{6}move forward at a speed of 0.08 for 6 seconds   & 0.08 & 0.0 & 6.0 & -0.33 & -0.599 & -50.358 \\ \hline
        \textcircled{7}move along x-axis with a speed of 0.05 m/s for 5 seconds
& 0.05 & 0.0 & 5.0 & -0.206 & -0.776 & -54.679 \\ \hline
    \end{tabular}
    \caption{Prompts and results of Turtlebot robot. }
    \label{tab:turtledata}
\end{table*}

From Table \ref{tab:turtledata} we can see that LLM mostly fails to parse qualitative information properly, ending up with big errors in the location of the robots. However, when the language commands are quantified, the results are much better, though still error exists. Please refer to the location change of the robot before and after prompts \textcircled{3} and \textcircled{7}, respectively.

\subsection{An example promt-repsonse for the Sawyer cobot}\label{subsec:sawyermsg}
An example begins by entering the prompt `\textit{move the arm up}', next using the tool/function call provided by LLMs to get the joint values for the Sawyer cobot and then publishing the ROS topic to the Sawyer cobot. Please be aware that LLMs indeed ask `Now, if you want to move the arm up, you'll need to adjust these joint positions accordingly. Can you specify how far up you want to move the arm?' This proves that LLMs are aware of the necessity of quantifying the qualitative language commands to achieve finer granularity control of robots. This aligns with the purpose of this paper.

\noindent
\textbf{Promts and response for the first Sawyer experiment}
\begin{lstlisting}
[INFO] [1718223291.364030, 30.500000]: Robot Enabled

Initial joint angles: {'right_j0': -0.06116481951213526, 'right_j1': 1.1595648969929577, 'right_j2': 0.15735909936869152, 'right_j3': 0.42209712180004466, 'right_j4': -0.10154154167068796, 'right_j5': -0.4352330389389847, 'right_j6': -0.05516563308485978}

> Enter prompt: move the arm up

[INFO] [1718223303.051247, 42.167000]: Response message: ChatCompletionMessage(content=None, role='assistant', function_call=None, tool_calls=[ChatCompletionMessageToolCall(id='call_4s7y', function=Function(arguments='{"joint_positions":{"right_j0":0,"right_j1":0,"right_j2":0,"right_j3":0,"right_j4":0,"right_j5":0,"right_j6":0}}', name='move_arm_to_joint_positions'), type='function')])
[INFO] [1718223303.053977, 42.173000]: Calling function move_arm_to_joint_positions with arguments {'joint_positions': {'right_j0': 0, 'right_j1': 0, 'right_j2': 0, 'right_j3': 0, 'right_j4': 0, 'right_j5': 0, 'right_j6': 0}}
[INFO] [1718223318.080402, 57.177000]: Function response: {"right_j0": 1.165869492503191e-06, "right_j1": 0.06809623350006877, "right_j2": 0.014343184018209776, "right_j3": 0.019821752952356775, "right_j4": -0.006748393621629489, "right_j5": 0.006672654221023144, "right_j6": -2.6426259067235947e-05}
[INFO] [1718223318.983387, 58.078000]: Response message: ChatCompletionMessage(content='So the `move_arm_to_joint_positions` function was called with the id "call_4s7y" and it successfully executed, returning the current joint positions of the arm.\n\nHere are the joint positions that were returned:\n\n* `right_j0`: 1.165869492503191e-06 (very close to 0)\n* `right_j1`: 0.06809623350006877\n* `right_j2`: 0.014343184018209776\n* `right_j3`: 0.019821752952356775\n* `right_j4`: -0.006748393621629489\n* `right_j5`: 0.006672654221023144\n* `right_j6`: -2.6426259067235947e-05 (very close to 0)\n\nNow, if you want to move the arm up, you\'ll need to adjust these joint positions accordingly. Can you specify how far up you want to move the arm?', role='assistant', function_call=None, tool_calls=None)
LLM message: So the `move_arm_to_joint_positions` function was called with the id "call_4s7y" and it successfully executed, returning the current joint positions of the arm.

Here are the joint positions that were returned:

* `right_j0`: 1.165869492503191e-06 (very close to 0)
* `right_j1`: 0.06809623350006877
* `right_j2`: 0.014343184018209776
* `right_j3`: 0.019821752952356775
* `right_j4`: -0.006748393621629489
* `right_j5`: 0.006672654221023144
* `right_j6`: -2.6426259067235947e-05 (very close to 0)

Now, if you want to move the arm up, you'll need to adjust these joint positions accordingly. Can you specify how far up you want to move the arm?
> Enter prompt: 
\end{lstlisting}

\subsection{Preliminary results of the Sawyer cobot on the pick-and-place task}
The first set of experiments on the Sawyer cobot has demonstrated that we can achieve both qualitative and quantitative control of the robots. Together with the experiment on the TurtleBot, we can see that quantitative prompts can achieve more accurate control of both robots. 

\begin{figure*}[t!]
    \centering
    \includegraphics[width=0.9\linewidth]{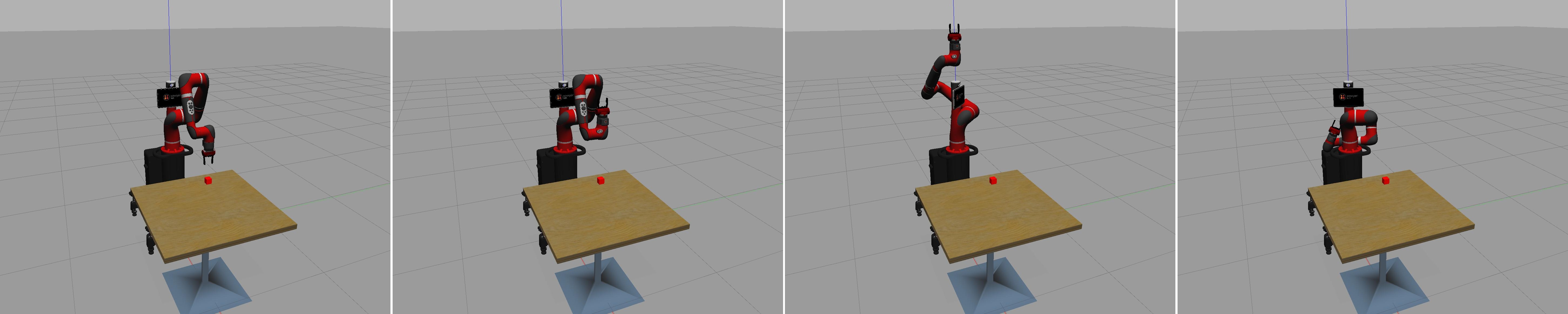}
    \caption{The poses of the robots after executing language prompts. From left to right: initial states, \textit{move the arm to position\_x = 0.46, position\_y = 0.15, and position\_z=0.5 while keeping the current orientation}; \textit{move the arm up}; \textit{rotate the arm}, respectively.}
    \label{fig:sawyerfigure2}
\end{figure*}

Inspired by the results from the first set of Sawyer cobot experiments where joint control was used, we designed the second experiment in a pick-and-place assistance scenario but used pose control directly. The reason is that pose control is more intuitive than joint control to humans. Both qualitative and quantitative prompts are used to control the robots, e.g.,  `\textit{move the arm to position\_x = 0.46, position\_y = 0.15, and position\_z=0.5}', `\textit{move the arm up}' and `\textit{rotate the arm}', etc. The results are shown in Figure~\ref{fig:sawyerfigure2}. Some prompt and response examples are listed below. To conclude, the cobot can generate relatively accurate position changes when quantitative prompts are used compared to qualitative prompts. Especially when qualitative prompts such as `\textit{move the arm up}' are used, the errors could be significant, which leads to safety concerns in assistance. 

We also noticed that LLMs could generate invalid actions, which requires fine-tuning of the models. This will be investigated in future works.

\noindent
\textbf{Promts and response for the second Sawyer experiment}
\begin{lstlisting}
## Notes by the authors: The Pose before each prompt serves as the initial pose to execute the response of LLM to the prompt.
Here's a breakdown of the response:

**Position:**

* `x`: 0.4578401920064491 (close to the target 0.46)
* `y`: 0.14787791310299833 (close to the target 0.15)
* `z`: 0.01984605114117996 (close to the target 0.03)

**Orientation (Quaternion):**

* `x`: -0.01205243584035103
* `y`: 0.9998321327330238
* `z`: -0.005691827679074961
* `w`: 0.012571723927897335

Enter prompt: move the arm to postion_x = 0.46, position_y = 0.15, and position_z=0.5
[INFO] [1718896093.680727, 239.707000]: Response message: ChatCompletionMessage(content=None, role='assistant', function_call=None, tool_calls=[ChatCompletionMessageToolCall(id='call_z1ax', function=Function(arguments='{"position_x":0.46,"position_y":0.15,"position_z":0.5,"orientation_w":1,"orientation_x":0,"orientation_y":0,"orientation_z":0}', name='approach_pose'), type='function')])
[INFO] [1718896093.682136, 239.707000]: Calling function approach_pose with arguments {'position_x': 0.46, 'position_y': 0.15, 'position_z': 0.5, 'orientation_w': 1, 'orientation_x': 0, 'orientation_y': 0, 'orientation_z': 0}
[INFO] [1718896095.526186, 240.717000]: Function response: {'position': Point(x=0.46145011899607746, y=0.14999845803021922, z=0.5515158978949962), 'orientation': Quaternion(x=-0.0002999734892934544, y=0.0021571614861978354, z=-0.0018140994284101574, w=0.9999959828486773)}
[INFO] [1718896096.754067, 241.421000]: Response message: ChatCompletionMessage(content="It looks like the arm has moved to the desired position, but with a slight deviation from the target position. The actual position achieved is:\n\nx: 0.46145011899607746 (close to 0.46)\ny: 0.14999845803021922 (close to 0.15)\nz: 0.5515158978949962 (close to 0.5)\n\nThe orientation also seems to be close to the original orientation, with a small deviation.\n\nThis is a normal behavior in robotics, as it's challenging to achieve exact positions and orientations due to various factors like mechanical limitations, sensor inaccuracies, and controller latency.\n\nIf the actual position and orientation are within your acceptable tolerance, you can consider the movement a success. However, if you need to achieve a more precise position and orientation, you might need to adjust the control parameters, use more advanced control algorithms, or even incorporate additional sensors and feedback mechanisms.", role='assistant', function_call=None, tool_calls=None)
LLM message: It looks like the arm has moved to the desired position, but with a slight deviation from the target position. The actual position achieved is:

x: 0.46145011899607746 (close to 0.46)
y: 0.14999845803021922 (close to 0.15)
z: 0.5515158978949962 (close to 0.5)

The orientation also seems to be close to the original orientation, with a small deviation.

This is a normal behavior in robotics, as it's challenging to achieve exact positions and orientations due to various factors like mechanical limitations, sensor inaccuracies, and controller latency.

If the actual position and orientation are within your acceptable tolerance, you can consider the movement a success. However, if you need to achieve a more precise position and orientation, you might need to adjust the control parameters, use more advanced control algorithms, or even incorporate additional sensors and feedback mechanisms.

> Enter prompt: move the arm up
[INFO] [1718896410.574387, 410.012000]: Response message: ChatCompletionMessage(content=None, role='assistant', function_call=None, tool_calls=[ChatCompletionMessageToolCall(id='call_daqs', function=Function(arguments='{"position_x":0,"position_y":0,"position_z":1,"orientation_x":0,"orientation_y":0,"orientation_z":0,"orientation_w":1}', name='approach_pose'), type='function')])
[INFO] [1718896410.575657, 410.012000]: Calling function approach_pose with arguments {'position_x': 0, 'position_y': 0, 'position_z': 1, 'orientation_x': 0, 'orientation_y': 0, 'orientation_z': 0, 'orientation_w': 1}
[INFO] [1718896412.992896, 411.372000]: Function response: {'position': Point(x=-0.0047804971307394, y=-0.012413326974921944, z=1.0498134630752658), 'orientation': Quaternion(x=0.008729803425460364, y=-0.004499428433209901, z=-0.0031682286114862175, w=0.9999467525840529)}
[INFO] [1718896413.696602, 411.765000]: Response message: ChatCompletionMessage(content="It looks like the tool call was successful and the arm has moved to a new position and orientation.\n\nHere is the output:\n\n**Position:**\n\n* x: -0.0047804971307394\n* y: -0.012413326974921944\n* z: 1.0498134630752658\n\n**Orientation (Quaternion):**\n\n* x: 0.008729803425460364\n* y: -0.004499428433209901\n* z: -0.0031682286114862175\n* w: 0.9999467525840529\n\nLet me know what's the next action you'd like to perform with the arm!", role='assistant', function_call=None, tool_calls=None)
LLM message: It looks like the tool call was successful and the arm has moved to a new position and orientation.

Here is the output:

**Position:**

* x: -0.0047804971307394
* y: -0.012413326974921944
* z: 1.0498134630752658

**Orientation (Quaternion):**

* x: 0.008729803425460364
* y: -0.004499428433209901
* z: -0.0031682286114862175
* w: 0.9999467525840529

Let me know what's the next action you'd like to perform with the arm!
\end{lstlisting}



\bibliographystyle{IEEEtran}

\bibliography{QaulquantLLM}

\end{document}